\pdfoutput=1

\documentclass[11pt]{article}

\usepackage{afterpage}
\usepackage{amsmath}
\usepackage{makecell}
\usepackage{arydshln}
\usepackage{geometry}
\usepackage{changepage} 

\usepackage{acl}
\usepackage{times}
\usepackage{latexsym}
\usepackage{booktabs}
\usepackage{caption}
\usepackage{tabularx}

\usepackage{tcolorbox}
\usepackage{lipsum}
\usepackage{listings}

\usepackage{amssymb}
\usepackage{pifont}
\newcommand{\cmark}{\ding{51}}%
\newcommand{\xmark}{\ding{55}}%

\usepackage[noabbrev,capitalize]{cleveref}

\def\oossymb{$\wr$}
\def\oosmark{\llap{\oossymb{}\,}}
\def\max#1{\textbf{#1}}

\def\parcite#1{\citep{#1}}  
\def\perscite#1{\citet{#1}}  
\def\inparcite#1{\citealp{#1}}  

\usepackage[T1]{fontenc}

\usepackage[utf8]{inputenc}

\usepackage{microtype}

\usepackage{inconsolata}
\usepackage{float}

\def\footurl#1{\footnote{\url{#1}}}

\usepackage[normalem]{ulem} 
\usepackage{xcolor}

\usepackage{graphicx}

\title{Findings of the Third Automatic Minuting (AutoMin) Challenge}

\author{Kartik Shinde \\
  Independent AI Researcher \\
  \small\texttt{kartikkts46@gmail.com} \\\And
  Laurent Besacier \\
  NAVER Labs Europe \\
  Grenoble, France \\
  \small\texttt{laurent.besacier@naverlabs.com} 
  \\\And
  Ond\v{r}ej Bojar \\
  Charles University, MFF, \'{U}FAL \\
  Prague, Czech Republic \\
  \small\texttt{bojar@ufal.mff.cuni.cz}
  \\\AND
  Thibaut Thonet \\
  NAVER Labs Europe \\
  Grenoble, France \\
  \small\texttt{thibaut.thonet@naverlabs.com} \\\And
  Tirthankar Ghosal \\
  Oak Ridge National Laboratory \\
  Oak Ridge, TN, USA \\
  \small\texttt{ghosalt@ornl.gov}
  \\
  }

\def\kmjec{Team Iterate}

\begin{document}
\maketitle
\begin{abstract}
This paper presents the third edition of AutoMin, a shared task on automatic meeting summarization into minutes. In 2025, AutoMin featured the main task of minuting—the creation of structured meeting minutes—as well as a new task: question answering (QA) based on meeting transcripts.

The minuting task covered two languages, English and Czech, and two domains: project meetings and European Parliament sessions. The QA task focused solely on project meetings and was available in two settings: monolingual QA in English, and cross-lingual QA, where questions were asked and answered in Czech based on English meetings.

Participation in 2025 was more limited compared to previous years, with only one team joining the minuting task and two teams participating in QA. However, as organizers, we included multiple baseline systems to enable a comprehensive evaluation of current (2025) large language models (LLMs) on both tasks.
\end{abstract}

\section{Introduction}

We report on AutoMin 2025, the third instance of the bi-annual shared task on meeting summarization into \emph{structured meeting minutes} \citep{automin:2021,automin:2023}.\footnote{See \url{https://elitr.github.io/automatic-minuting} and \url{https://ufal.github.io/automin-2023/} for the previous instances and \url{https://ufal.github.io/automin-2025/} for the current one.}
We build upon our past experience \cite{muskaan-singh-bojar-2021-empirical, shinde21_automin} from two previous AutoMins and add a new challenge to facilitate personalized access to meetings: answering questions about meeting content.

As in the previous editions, the meeting summarization task is run separately in English and Czech. The new question-answering task is run monolingually (English-only) and cross-lingually (Czech questions about English meetings).

\section{Tasks Description}

AutoMin 2025 reuses the main task from the 2023 challenge and adds a new one.

\subsection{Task A: Minuting}

\textit{The main task consists of automatically generating minutes from multiparty meeting conversations} provided in the form of transcripts. The objective is to generate minutes as bulleted lists, summarizing the main contents of the meeting, as opposed to usual paragraph-like text summaries \cite{shinde-etal-2022-automatic}.

As in 2023 \cite{ghosal-etal-2022-second}, Task A is run in two domains. In English and Czech, we again
rely on the meetings in the ELITR Minuting Corpus 1.0
(denoted ELMI for short; \inparcite{nedoluzhko-etal-2022-elitr}).
For English, we again add EuroParlMin 1.0,\footurl{https://github.com/ufal/europarlmin} a resource we curated from the European Parliament sessions for AutoMin 2023.

Participants were free to submit their minutes for any selection of these test sets. 
Note that the nature of meetings as well as the reference minutes are very different in the two datasets (technical project meetings vs. parliamentary sessions).


%
%
%
%

\subsection{Task B: Meeting QA}

\textit{Task B is new and consists in answering questions based on meeting transcripts.}

\def\ebench{{ELITR\discretionary{-}{}{-}Bench}}
\def\ebenchqa{{ELITR\discretionary{-}{}{-}Bench\discretionary{-}{}{-}QA}}
\def\ebenchconv{{ELITR\discretionary{-}{}{-}Bench\discretionary{-}{}{-}Conv}}

The evaluation builds on the work of \citet{thonet2024elitrbenchmeetingassistantbenchmark} who recently created \ebench{}, a collection of
271 manually crafted questions in English along with their corresponding ground-truth answers related to a subset of English meeting minutes from the ELITR Minuting Corpus. Subsequently, \ebench{} (i.e. the set of questions and reference answers, not the underlying English meeting transcripts) was manually translated to Czech. 

In this challenge, the participants were asked to answer these questions either monolingually (English questions on English transcripts) or cross-lingually (Czech questions on English transcripts), using ASR-generated meeting transcripts.

Given that meetings tend to be lengthy (with an average one-hour meeting generating around 20,000 tokens), solving this task effectively requires large language models capable of handling long contexts.

\section{Evaluation}

\subsection{Task A: Minuting}

In Task A, we are interested in four different criteria for the minutes. We adopt the same scales as used in the previous edition, see \perscite{automin:2023}.

\begin{enumerate}
    \item \textbf{Adequacy} assesses if the minutes adequately capture the major topics discussed in the meeting, also considering coverage (all such topics covered).
    \item \textbf{Fluency} reflects if the minutes consist of fluent, coherent texts and are readable to the evaluator.
    \item \textbf{Grammatical Correctness} checks the level to which the minutes are grammatically correct.
    \item \textbf{Relevance} signifies the extent to which the minutes overall capture the important content from the source transcript (as opposed to summarizing useless parts).
\end{enumerate}

Short of funding for manual evaluation of submitted outputs, we evaluate Task A only automatically.

Our primary evaluation is LLM-based (see \cref{sec:gpt_eval}) and we also complement it with classical automatic evaluation measures like ROUGE (see \cref{sec:auto_eval}).

\subsubsection{Automatic Evaluation with LLMs}
\label{sec:gpt_eval}

To obtain estimates for meeting minutes on the different quality scales, we relied on the capabilities of large language models (LLMs), particularly GPT-4 \cite{achiam2023gpt}.

Importantly, we build upon our experience from the 2023 AutoMin edition, see Sections~7.2 and 8 in \perscite{automin:2023}. The manual evaluation at the ``chunk'' level (i.e. the level of individual items in the minutes) was found less adequate than the full document-level manual evaluation. GPT-4-based evaluation was not sensitive to the different scales (Adequacy, Fluency, Grammaticality and Relevance): all its scorings were pairwise correlated over .8 Pearson. The agreement of GPT-4 scores and manual document-level scores was poor. The ``pairwise accuracy'' (a simplification of Kendall's Tau, see \inparcite{kocmi-etal-2021-ship}), i.e. the proportion of system pairs where both GPT and humans order the pair equally, was always under 2/3 and often close to the chance level of .5 across the different scales and two prompt settings. (ROUGE-1 and ROUGE-L, on the other hand, easily surpassed the pairwise accuracy of 2/3.)

In order to improve chances for a more reliable GPT-based evaluation this year, we employed a structured prompting framework inspired by chain-of-thought (CoT)-style reasoning, tailored to summarization assessment. Specifically, we instructed a high-capacity language model to perform detailed, step-by-step evaluations.
The full prompt is available in \ref{gpt2025-prompt} of the Appendix.

\citet{wei2022chain} demonstrate that prompting large language models to ``think step-by-step'' leads to noticeable improvements in complex reasoning tasks. 
We wish to validate this observation in our case.

Specifically, the prompt was designed to guide the LLM through a two-step reasoning process.

\begin{itemize}
    \item \textbf{Step 1: Information Mapping.} The model was asked to extract and list key information units from the summary and link each to supporting evidence from the transcript. This step hopefully enforced a grounding mechanism between the output and the input data.
    
    \item \textbf{Step 2: Metric-Specific Scoring.} For each of the four evaluation criteria (\textit{Adequacy}, \textit{Fluency}, \textit{Grammaticality}, and \textit{Relevance}) the model was asked to provide a brief justification and a score on a Likert Scale of 1--5, grounded in the original definitions sourced from prior literature.
\end{itemize}

After the evaluation (still within the same prompt), the model was asked to perform a lightweight self-consistency check: the model was explicitly instructed to revisit the lowest assigned score and verify whether it could be reasonably increased based on the definitions. We hope this approach helped reduce stochastic variance and promoted more reliable factual judgments. Overall, the prompting strategy was intended to enhance interpretability, ensure grounding in the data, and improve factual alignment in scoring.

Finally, the prompt requested the model to produce outputs in a particular JSON format to simplify results' extraction.

To run the prompt, we used a deterministic generation configuration (temperature = 0, top-p = 1.0), ensuring scoring consistency and minimizing output variance.

We observed that in all cases, the final output adhered strictly to the requested schema, allowing easy aggregation of the scores. By combining structured CoT prompting, definition-based anchoring, and minimal generation entropy, we hope that this protocol enabled scalability and offered a reproducible surrogate for manual scoring.

\subsubsection{Automatic Evaluation with Standard Metrics}
\label{sec:auto_eval}

For the ``standard'' automatic evaluation of Task A submissions, we rely on a combination of n-gram and embedding-based metrics to comprehensively assess the quality of the generated summaries. These metrics included the ROUGE family (ROUGE-1, ROUGE-2, ROUGE-L), as well as BERTScore and BARTScore. While automatic evaluation remains a noisy proxy for summary quality in the context of meeting transcripts, these metrics offer useful signals when interpreted with care.

\paragraph{ROUGE Variants.} ROUGE (Recall-Oriented Understudy for Gisting Evaluation) \cite{lin2004rouge} is a standard metric in summarization, based on measuring lexical overlap between a candidate and reference summary. We use the F1 variants of ROUGE-1, ROUGE-2, and ROUGE-L:

\begin{itemize}
    \item \textbf{ROUGE-1:} Overlap of unigrams between candidate and reference.
    \item \textbf{ROUGE-2:} Overlap of bigrams.
    \item \textbf{ROUGE-L:} Measures the length of the longest common subsequence (LCS) between candidate and reference, capturing fluency and sentence-level alignment.
\end{itemize}

While often interpreted as recall-oriented, ROUGE can also be computed in terms of precision. The recall and precision definitions used in ROUGE are:

\begin{adjustwidth}{-1em}{}
\begin{equation}
\text{ROUGE}_{\text{Recall}} =
  \frac{\#\,\text{Overlapping n-grams}}
       {\substack{\text{Total n-grams in}\\\text{Reference Summary}}}
\end{equation}
\end{adjustwidth}

\begin{adjustwidth}{-1em}{}
\begin{equation}
\text{ROUGE}_{\text{Precision}} =
  \frac{\#\,\text{Overlapping n-grams}}
       {\substack{\text{Total n-grams in}\\\text{Candidate Summary}}}
\end{equation}
\end{adjustwidth}

In our evaluation, we used the ROUGE F1 score across all ROUGE variants, as it represents the harmonic mean of precision and recall.

\paragraph{BERTScore.} BERTScore \cite{zhang2019bertscore} evaluates the semantic similarity between generated and reference summaries using contextual embeddings from BERT. It computes:

\begin{itemize}
    \item \textbf{Precision:} Average cosine similarity between each token in the candidate and its most similar token in the reference.
    \item \textbf{Recall:} Average cosine similarity between each token in the reference and its most similar token in the candidate.
    \item \textbf{F1:} Harmonic mean of precision and recall.
\end{itemize}

Unlike ROUGE, BERTScore is robust to synonymy and paraphrasing, making it better suited for abstractive summarization evaluation where wording may vary substantially from the reference. We used the Python standard implementation\footnote{\url{https://pypi.org/project/bert-score/}} of BERTScore.

\paragraph{BARTScore.} BARTScore \cite{NEURIPS2021_e4d2b6e6} reframes summary evaluation as a language modeling problem using a pretrained sequence-to-sequence model. The core idea is that the likelihood of generating the reference from the summary (or vice versa) correlates with its quality. We compute BARTScore in the reference-conditioned direction using the following formulation:

\begin{adjustwidth}{-1em}{}
\begin{equation}
\text{BARTScore} = \sum_{t=1}^{m} \omega_t \cdot \log p(y_t \mid y_{<t}, x, \theta)
\end{equation}
\end{adjustwidth}

Here, $y$ is the generated summary, $x$ is the reference, $\omega_t$ is a positional weight (typically uniform), and $\theta$ are the model parameters. This approach captures both fluency and semantic alignment and has gained popularity for abstractive generation tasks.\footnote{Since BARTScore is the sum of log-probabilities, scores are typically negative, and less negative (i.e., higher) values indicate better alignment and fluency.}
We use the original implementation from the authors.\footnote{\url{https://github.com/neulab/BARTScore}} 

Overall, these automatic metrics offer complementary perspectives: ROUGE captures surface-level matching, BERTScore captures semantic similarity, and BARTScore evaluates generation quality from a probabilistic modeling standpoint.

%
%

\subsection{Task B: Meeting QA}

The evaluation of question answering on meeting transcripts adheres to the methodology introduced in \citet{thonet2024elitrbenchmeetingassistantbenchmark}. The approach is based on large language models as automated judges to assess the quality of responses. Specifically, these models compare the system-generated answers to the human-crafted gold reference answer for each given query. Following \citet{thonet2024elitrbenchmeetingassistantbenchmark,kim2023prometheus}, our LLM-as-a-judge is provided with a 10-point score rubric that details the expected quality criteria for each grade level in order to guide the LLM towards the relevant numeric score. The prompts that we used are based on the ones introduced in these latter works, and they are detailed in \cref{fig:task-b-prompt-mono,fig:task-b-prompt-cross} in the Appendix for the monolingual task and the cross-lingual task, respectively.

The LLM adopted for the evaluation is GPT-4o.\footnote{We specifically used the gpt-4o-2024-11-20 model.} This choice is motivated by the fact that GPT-4 was shown to be highly correlated with human evaluators in \citet{thonet2024elitrbenchmeetingassistantbenchmark}.


\begin{table}[H]
\centering
\resizebox{0.48\textwidth}{!}{%
\begin{tabular}{lcc}
\hline
 & \textbf{Lines} & \textbf{Words} \\
\hline
\multicolumn{3}{l}{\textbf{ELMI}} \\
\hline
Transcripts & 748.6$\pm$305.8 & 6554.6$\pm$2542.0 \\
Ref. Minutes & 45.3$\pm$18.6 & 429.0$\pm$246.2 \\
\hline
Darbarer & 35.8$\pm$14.1 & 290.6$\pm$142.3 \\
davinci003 & 40.6$\pm$12.6 & 451.1$\pm$180.2 \\
text-davinci-003 & 32.6$\pm$9.6 & 484.2$\pm$144.9 \\
GPT-4 & 34.6$\pm$13.3 & 495.4$\pm$212.0 \\
NTR & 83.1$\pm$37.3 & 854.2$\pm$374.8 \\
Synapse & 35.3$\pm$10.5 & 404.8$\pm$152.8 \\
KMJEC & 32.4$\pm$13.8 & 379.9$\pm$156.4 \\
Zoom-long & 34.1$\pm$9.0 & 561.8$\pm$146.5 \\
Zoom-short & 6.2$\pm$1.3 & 107.1$\pm$23.8 \\
GPT-4 (2025) & 21.1$\pm$9.7 & 387.8$\pm$190.7 \\
\textit{Hallucination Index:} & & \\
\quad distilbart & 6.7$\pm$2.6 & 52.2$\pm$12.2 \\
\quad bart-samsum & 3.2$\pm$1.0 & 35.9$\pm$6.9 \\
\quad pegasus-xsum & 1.2$\pm$0.6 & 78.4$\pm$58.2 \\
\quad t5-small & 4.9$\pm$3.1 & 45.2$\pm$13.2 \\
\hline
\multicolumn{3}{l}{\textbf{EuroParlMin}} \\
\hline
Transcripts & 55.9$\pm$112.0 & 851.1$\pm$2044.6 \\
Ref. Minutes & 27.1$\pm$47.1 & 167.6$\pm$291.9 \\
\hline
Darbarer & 6.7$\pm$4.4 & 58.4$\pm$68.5 \\
davinci-003 & 7.1$\pm$3.0 & 101.4$\pm$49.0 \\
GPT-4 & 5.2$\pm$0.9 & 78.0$\pm$26.4 \\
NTR\_NLP\_Team & 9.9$\pm$19.0 & 149.7$\pm$301.7 \\
Team Synapse & 10.0$\pm$10.5 & 107.5$\pm$174.0 \\
GPT-4 (2025) & 9.9$\pm$7.4 & 198.4$\pm$233.6 \\
\textit{Hallucination Index:} & & \\
\quad distilbart & 2.7$\pm$1.1 & 52.3$\pm$10.5 \\
\quad bart-samsum & 2.7$\pm$0.8 & 47.1$\pm$9.6 \\
\quad pegasus-xsum & 1.3$\pm$0.5 & 56.4$\pm$25.3 \\
\quad t5-small & 2.8$\pm$0.8 & 46.5$\pm$9.6 \\
\hline
\end{tabular}}
\caption{Basic properties of manual transcripts, reference minutes and all team submissions of test set meetings. Hallucination Index models are grouped for compactness. We report average$\pm$standard deviation values.}
\label{tab:lines_and_word_counts}
\end{table}

\section{Datasets}


We provided AutoMin 2025 participants with ELITR Minuting Corpus \citep{nedoluzhko-etal-2022-elitr} and we also allowed them to use any external dataset if they explicitly describe them in their system reports.

\subsection{ELITR Minuting Corpus for Task A}

\begin{table}[t]
    \centering
\small
\begin{tabular}{l@{~~}r@{~~}r@{~~}r@{~~}r}
\hline
 & \multicolumn{2}{c}{English} & \multicolumn{2}{c}{Czech} \\\cline{2-5}
Meeting Minuted & \#meetings & \#hours & \#meetings & \#hours \\
\hline
Once & 30 & 22 & 8 & 2 \\
Twice & 65 & 65 & 20 & 20 \\
More than twice & 25 & 22 & 31 & 31 \\
\hline
\textbf{Total meetings} & 120 & 109 & 59 &  53 \\\hline
\end{tabular}
  \caption{Basic transcript and minutes statistics for ELITR Minuting Corpus.}
    \label{tab:ourdataset_statistics}
\end{table}

In our ELITR Minuting Corpus \citep{nedoluzhko-etal-2022-elitr}, a meeting usually contains one manually corrected transcript, one original minute (created by a meeting participant; in some cases, these minutes are a detailed agenda which got further updated during or after the meeting), and one or more minutes created later by annotators who were not present at the meeting themselves (denoted `generated' in the corpus).

\cref{tab:ourdataset_statistics} presents our dataset's statistics regarding the number of meetings and hours. We separately count meetings for which we have only one, two, and more than two (up to 11) independently created minutes. For English meetings, either (i)  our annotators created both minutes or (ii) one minute was written by one of the participants before or after the meeting and another by our annotator. In contrast, most meetings in the Czech portion of the dataset are minuted at least twice, and more than half of the Czech portion of ELITR Minuting Corpus is minuted 3-5 times.

\textit{Note that for AutoMin 2025, only the English portion of the ELITR Minuting Corpus was used.}

To address GDPR requirements and protect the privacy of meeting participants, we de-identify any reference concerning specific persons, organisations, projects and locations. Such names were replaced with the lexical substitute strings [PERSON\textit{number}], [ORGANIZATION\textit{number}], [PROJECT\textit{number}] and [LOCATION\textit{number}], respectively. The square brackets indicate a mention of that entity. The corpus also has the strings  (PERSON\textit{number}) at the beginning of nearly every line which denote the speaker of the utterance. Additionally, we replaced the names of annotators mentioned in minutes with [ANNOTATOR\textit{number}].

\subsection{EuroParlMin for Task A}

We curated this dataset from the publicly available European parliamentary sessions by using the transcripts in the EuroParl dataset \citep{koehn:europarl:mtsummit:2005} and crawling the corresponding minutes from the EU parliament website.\footnote{\url{https://www.europarl.europa.eu/committees/en/meetings/minutes}}

We automatically created a set of transcript--minute pairs ($\sim$2000) for AutoMin 2023 and we reused it in 2025.

\subsection{ELITR-Bench for Task B}

Task B leverages \ebench{}\footnote{\url{https://github.com/utter-project/ELITR-Bench}} – a benchmark for the evaluation of long-context LLMs on meeting transcripts. The meeting data used in this benchmark originally come from the ELITR Minuting Corpus.

\begin{figure}[t]
    \centering
\fbox{
\begin{minipage}{0.9\columnwidth}
\small
Q: Who were the participants of the meeting? \\
A: [PERSON14], [PERSON10], [PERSON5], [PERSON9], [PERSON1], [PERSON11] \\
\\
Q: What was the main purpose of this meeting? \\ 
A: Discuss and finalize the technical setup for a demo \\
\\
Q (Conv version): How many scenarios were discussed? \\
Q (QA version): How many scenarios were discussed \uline{for the demo setup}? \\
A: 3 (plans A, B and C) \\
\\
Q (Conv version): Which scenario was chosen eventually? \\
Q (QA version): Which scenario was chosen eventually \uline{for the demo setup}? \\
A: Plan C
\end{minipage}
}
    \caption{An illustration of variants of \ebench{} questions for the QA and Conv versions. No mention of the version indicates that the same question is used for both versions. Differences underlined for presentation purposes.}
    \label{elitr-bench-questions}
\end{figure}

For the cross-lingual variant, we used \ebench{} questions which we had professionally translated from English into Czech. This variant of the test set allows to evaluate the following setting: asking questions in Czech, locating answers in English meeting minutes, reporting answers in Czech and comparing them to the golden-truth Czech answer.

\ebench{} is available in two settings. In \ebenchqa{}, we designed for each meeting a set of stand-alone questions (along with their answers) that can be addressed solely based on the meeting transcript, without additional context. We also designed a modified \ebenchconv{} version where questions are to be asked in sequence, in a pre-defined order within a conversation. In this setting, some of the questions contain pronominal references or ellipses, for which previous conversational context (i.e., previous questions and answers) must be used to answer properly (see \cref{elitr-bench-questions}).
The two settings overlap greatly: of the total of 271 questions only 33 differ between the -QA and -Conv variant.

\textit{Note that for AutoMin 2025, only the -QA setting was used.}



\section{Submissions}

The participation in AutoMin 2025 was unfortunately much lower than in the previous years.

\subsection{Task A Submissions}

For Task A, we received one submission that focused only on the English
datasets, i.e. ELMI EN and EuroParlMin. To provide a better reference, we also
include the 2023 submissions in our evaluation and a baseline system using
GPT.

The sole 2025 submission is briefly described below, for the 2023 systems, we refer the reader to \perscite{automin:2023}.

\paragraph{HallucinationIndexes \rm\parcite{katwe2025}}
introduces a reinforcement learning-based approach to reduce hallucinations in summarization. They propose the Entity Hallucination Index (EHI)
as a reward signal, which measures factual alignment between generated summaries
and source documents using named entities. The summarization model, based on
BART, is fine-tuned with this EHI-based reward to penalize entity-level
hallucinations. Unlike teams from the previous iteration of AutoMin that focus
on segmentation and summarization pipelines, the HallucinationIndexes team directly
targets the factual correctness of the output, making their method orthogonal
and potentially complementary to structure-focused pipelines like those of team
Iterate or Darbarer in 2023.

In addition, their approach does not rely on meeting-specific preprocessing steps such as speaker-aware segmentation or dialogue structuring. Instead, they focus on a model-centric strategy that is domain-agnostic and applicable to any summarization task where factual consistency is critical. The use of reinforcement learning also sets their method apart, as none of the previous AutoMin participants explored training summarization models via reward-driven optimization.

\paragraph{GPT-2025} are our baseline minutes obtained using
GPT-4 and the same prompt as we used for
GPT baseline in 2023, namely: 
``Summarize the following project meeting in the form of 5 - 10 bullet points: \textless meeting transcript\textgreater''.

\subsection{Task B Submissions}

The submissions for Task B consist of two systems which are detailed below. In
addition to these systems, we include a comparison against
baselines that simply involve prompting an LLM with the meeting transcript in
its context, following the same approach as described
in \citet{thonet2024elitrbenchmeetingassistantbenchmark}. The LLMs considered in
these baselines are the following: GPT-4o,\footnote{We used the gpt-4o-2024-11-20
model.}
LLaMA-3.2-3B-Instruct,\footnote{\url{https://huggingface.co/meta-llama/Llama-3.2-3B-Instruct}}
LLaMA-3.1-8B-Instruct,\footnote{\url{https://huggingface.co/meta-llama/Llama-3.1-8B-Instruct}}
Phi-4-Mini-Instruct,\footnote{\url{https://huggingface.co/microsoft/Phi-4-mini-instruct}}
Phi-3-Small-128k-Instruct.\footnote{\url{https://huggingface.co/microsoft/Phi-3-small-128k-instruct}}
The prompts used for these baselines are detailed in the Appendix,
in~\cref{fig:answer-prompt-en,fig:answer-prompt-encz} for the monolingual and
cross-lingual subtasks, respectively.

\subsubsection{GETALP}

Team GETALP's system~\citep{gonzalez-saez2025} investigates the combination of Retrieval Augmented Generation (RAG) and Abstract Meaning Representation (AMR) techniques. The former consists in retrieving relevant passages from the meeting transcript while the latter builds a graph representation of the semantic components of the retrieved sentences. This results in three systems: RAG-only (the context only contains retrieved passages), AMR-only (the context only contains the sentences from the graph representations) and RAG+AMR (the context contains both retrieved passages and graph-based sentences). These approaches are extractive, requiring the prompted LLM~-- a LLaMA-3.1-8B-Instruct model~-- to provide the 1-2 most relevant sentences from the context. The different systems are validated on the monolingual (English-only) and cross-lingual (English-Czech) tasks.

\subsubsection{HallucinationIndexes}

The HallucinationIndexes system \citep{katwe2025} is briefly described above in Task A.

The base model used in Task B is Flan-T5 which has a limited context window of 1024 tokens and may thus struggle to process \ebench{} lengthy transcripts.

\begin{table*}[t]
  \centering
  \begin{tabular*}{\textwidth}{@{\extracolsep{\fill}} l c c c c }
	& \multicolumn{4}{c}{\bf GPT-2025-CoT} \\
    & \textbf{Adeq.} & \textbf{Flu.} & \textbf{Gram.} & \textbf{Relev.} \\
    \cmidrule{1-5}
    \multicolumn{5}{@{}l}{\textbf{ELMI EN}} \\
    \cmidrule{1-5}
GPT-4           	& \max{4.00$\pm$0.43}	& \max{4.42$\pm$0.51}    	& \max{4.75$\pm$0.45}    	& \max{4.33$\pm$0.65} \\
text-davinci-003	& 3.25$\pm$0.62      	& 3.92$\pm$0.67          	& 4.00$\pm$0.74          	& 3.33$\pm$0.78 \\
davinci-003     	& 3.08$\pm$0.51      	& 3.75$\pm$0.45          	& \oosmark{4.08$\pm$0.29}	& 3.25$\pm$0.87 \\
Zoom-long       	& 2.42$\pm$0.51      	& 3.00$\pm$0.60          	& 3.25$\pm$0.75          	& 2.42$\pm$0.51 \\
Synapse         	& 2.33$\pm$0.49      	& \oosmark{3.25$\pm$0.45}	& 3.17$\pm$0.58          	& 2.42$\pm$0.51 \\
Darbarer        	& 2.25$\pm$0.45      	& \oosmark{3.50$\pm$0.52}	& \oosmark{3.83$\pm$0.58}	& 2.33$\pm$0.49 \\
Zoom-short      	& 2.17$\pm$0.39      	& 3.00$\pm$0.60          	& 3.67$\pm$0.49          	& 2.17$\pm$0.39 \\
\kmjec{}        	& 2.08$\pm$0.51      	& 2.50$\pm$0.52          	& 3.00$\pm$0.43          	& 2.08$\pm$0.51 \\
NTR             	& 2.00$\pm$0.00      	& 2.08$\pm$0.29          	& 2.17$\pm$0.39          	& 2.00$\pm$0.00 \\
    \addlinespace[4pt]
    \cdashline{1-5}
    \addlinespace[4pt]
GPT-4 (2025)                	& \max{3.67$\pm$0.49}	& \max{4.42$\pm$0.51}	& \max{4.75$\pm$0.45}    	& \max{4.00$\pm$0.60}	\\
HallucinationIndexes@ (2025)	\\                   	                     	                         	                     	
\quad bart-samsum           	& 1.08$\pm$0.29      	& 2.92$\pm$0.51      	& 3.50$\pm$0.67          	& 1.17$\pm$0.39      	\\
\quad distilbart            	& 1.08$\pm$0.29      	& 2.17$\pm$0.39      	& 2.25$\pm$0.45          	& 1.08$\pm$0.29      	\\
\quad pegasus-xsum          	& 1.00$\pm$0.00      	& 2.17$\pm$0.58      	& \oosmark{2.92$\pm$0.90}	& 1.00$\pm$0.00      	\\
\quad t5-small              	& 1.00$\pm$0.00      	& 2.08$\pm$0.51      	& 2.42$\pm$0.67          	& 1.00$\pm$0.00      	\\
    \cmidrule{1-5}
    \multicolumn{5}{@{}l}{\textbf{EuroParlMin}} \\
    \cmidrule{1-5}
Synapse    	& \max{3.40$\pm$0.80}	& \max{4.62$\pm$0.58}    	& \max{4.83$\pm$0.51}    	& \max{3.99$\pm$0.78} \\
NTR        	& 2.80$\pm$1.04      	& 3.80$\pm$1.01          	& 4.02$\pm$1.07          	& 3.10$\pm$1.22 \\
Darbarer   	& 2.43$\pm$0.69      	& \oosmark{4.29$\pm$0.64}	& \oosmark{4.85$\pm$0.44}	& 3.02$\pm$0.83 \\
GPT-4      	& 1.58$\pm$1.24      	& 3.90$\pm$0.68          	& 4.70$\pm$0.57          	& 1.64$\pm$1.33 \\
davinci-003	& 1.57$\pm$1.21      	& 3.84$\pm$0.65          	& 4.64$\pm$0.54          	& 1.64$\pm$1.32 \\
    \addlinespace[4pt]
    \cdashline{1-5}
    \addlinespace[4pt]
GPT-4 (2025)                	& \max{4.78$\pm$0.58}	& \max{4.95$\pm$0.27}    	& \max{4.98$\pm$0.26}    	& \max{4.84$\pm$0.51}	\\
HallucinationIndexes@ (2025)	                     	                         	                         	                     	\\
\quad bart-samsum           	& 2.96$\pm$0.99      	& 4.41$\pm$0.68          	& 4.63$\pm$0.76          	& 3.52$\pm$1.16      	\\
\quad distilbart            	& 2.73$\pm$0.96      	& 4.02$\pm$0.85          	& 4.32$\pm$1.00          	& 3.23$\pm$1.16      	\\
\quad t5-small              	& 2.67$\pm$0.98      	& 3.00$\pm$0.69          	& 3.09$\pm$0.87          	& 3.22$\pm$1.27      	\\
\quad pegasus-xsum          	& 1.54$\pm$0.63      	& \oosmark{3.31$\pm$1.03}	& \oosmark{4.02$\pm$1.21}	& 1.84$\pm$0.79      	\\
    \bottomrule
  \end{tabular*}
  \caption{GPT-2025-CoT evaluation results (mean$\pm$std) on the ELMI and EuroParlMin datasets as summarized by 2023 systems (upper parts) and 2025 systems (lower parts). 
  The scores are on the scale 1 (worst) to 5 (best).
  Within each group, systems are sorted by decreasing Adequacy, and Fluency in the case of a tie.
  ``\oossymb'' denotes cells where the value increases, breaking the ordering. 
  Boldfaced scores are the best in each group, or within the $\pm$ bounds of the best scores.
  }
  \label{tab:task-a-gpt-2025-cot}
\end{table*}

\begin{table*}[t]
\centering
\begin{tabularx}{\textwidth}{@{}lccccc@{}}
 & \textbf{ROUGE-1} & \textbf{ROUGE-2} & \textbf{ROUGE-L} & \textbf{BART-F1} & \textbf{BERT-F1} \\
\cmidrule{1-6}
\multicolumn{6}{@{}l}{\textbf{ELMI EN}} \\
\cmidrule{1-6}
GPT-4           	& \max{0.44$\pm$0.06}	& \max{0.11$\pm$0.04}          	& \max{0.20$\pm$0.03}          	& \max{-4.14$\pm$0.31}          	& \max{0.42$\pm$0.04}          	\\
Synapse         	& \max{0.43$\pm$0.06}	& \max{0.11$\pm$0.04}          	& \max{0.20$\pm$0.02}          	& \oosmark{\max{-4.25$\pm$0.26}}	& \max{0.41$\pm$0.04}          	\\
text-davinci-003	& \max{0.41$\pm$0.07}	& \max{0.10$\pm$0.02}          	& \max{0.19$\pm$0.02}          	& \max{-4.07$\pm$0.24}          	& 0.35$\pm$0.02                	\\
\kmjec{}        	& \max{0.41$\pm$0.08}	& \max{0.10$\pm$0.03}          	& \max{0.19$\pm$0.03}          	& \oosmark{\max{-4.18$\pm$0.28}}	& 0.35$\pm$0.03                	\\
Zoom-long       	& \max{0.41$\pm$0.08}	& \max{0.10$\pm$0.03}          	& \max{0.18$\pm$0.02}          	& \max{-4.10$\pm$0.21}          	& 0.35$\pm$0.02                	\\
davinci-003     	& \max{0.41$\pm$0.07}	& \max{0.09$\pm$0.03}          	& \max{0.17$\pm$0.03}          	& \oosmark{\max{-4.29$\pm$0.31}}	& \oosmark{\max{0.39$\pm$0.03}}	\\
Darbarer        	& \max{0.40$\pm$0.06}	& \oosmark{\max{0.10$\pm$0.03}}	& \oosmark{\max{0.19$\pm$0.03}}	& \oosmark{\max{-4.35$\pm$0.17}}	& 0.39$\pm$0.03                	\\
NTR             	& 0.37$\pm$0.10      	& \max{0.09$\pm$0.04}          	& 0.16$\pm$0.03                	& \oosmark{\max{-4.52$\pm$0.27}}	& 0.35$\pm$0.05                	\\
Zoom-short      	& 0.28$\pm$0.08      	& 0.06$\pm$0.02                	& 0.15$\pm$0.04                	& \max{-4.52$\pm$0.22}          	& 0.30$\pm$0.03                	\\
\addlinespace[4pt]
\cdashline{1-6}
\addlinespace[4pt]
GPT-4 (2025)               	& \max{0.42$\pm$0.07}	& \max{0.11$\pm$0.03}	& \max{0.19$\pm$0.03}	& \max{-3.82$\pm$0.35}    	& \max{0.38$\pm$0.04}    	\\
HallucinationIndexes (2025)	\\                   	                     	                     	                          	                         	
\quad pegasus-xsum         	& 0.10$\pm$0.04      	& 0.01$\pm$0.01      	& 0.07$\pm$0.02      	& -3.62$\pm$1.12          	& 0.06$\pm$0.05          	\\
\quad bart-samsum          	& 0.08$\pm$0.03      	& 0.01$\pm$0.01      	& 0.06$\pm$0.02      	& \oosmark{-4.34$\pm$0.32}	& \oosmark{0.15$\pm$0.08}	\\
\quad distilbart           	& 0.08$\pm$0.03      	& 0.01$\pm$0.01      	& 0.05$\pm$0.01      	& \oosmark{-4.93$\pm$0.30}	& 0.13$\pm$0.04          	\\
\quad t5-small             	& 0.07$\pm$0.03      	& 0.01$\pm$0.01      	& 0.05$\pm$0.01      	& \oosmark{-5.00$\pm$0.49}	& 0.12$\pm$0.05          	\\
\cmidrule{1-6}
\multicolumn{6}{@{}l}{\textbf{EuroParlMin}} \\
\cmidrule{1-6}
Darbarer   	& \max{0.27$\pm$0.10}	& \max{0.11$\pm$0.08}	& \max{0.18$\pm$0.08}	& \max{-4.76$\pm$0.63}    	& \max{0.28$\pm$0.07}          	\\
NTR        	& \max{0.27$\pm$0.11}	& \max{0.09$\pm$0.07}	& \max{0.17$\pm$0.07}	& \max{-4.53$\pm$0.65}    	& \max{0.20$\pm$0.10}          	\\
Synapse    	& \max{0.26$\pm$0.10}	& \max{0.08$\pm$0.07}	& \max{0.16$\pm$0.08}	& \oosmark{-4.79$\pm$0.64}	& \oosmark{\max{0.23$\pm$0.08}}	\\
davinci-003	& \max{0.21$\pm$0.09}	& \max{0.04$\pm$0.05}	& \max{0.14$\pm$0.06}	& \max{-4.66$\pm$0.56}    	& 0.16$\pm$0.09                	\\
GPT-4      	& \max{0.20$\pm$0.09}	& \max{0.04$\pm$0.05}	& \max{0.13$\pm$0.06}	& \oosmark{-4.71$\pm$0.57}	& 0.15$\pm$0.10                	\\
\addlinespace[4pt]
\cdashline{1-6}
\addlinespace[4pt]
GPT-4 (2025)	& \max{0.32$\pm$0.10}	& \max{0.13$\pm$0.06}	& \max{0.21$\pm$0.06}	& \max{-3.79$\pm$0.48}	& \max{0.29$\pm$0.09}	\\
HallucinationIndexes@ (2025)\\
\quad pegasus-xsum	& \max{0.27$\pm$0.11}	& \max{0.10$\pm$0.09}	& \max{0.18$\pm$0.09}	& -3.38$\pm$0.68          	& 0.19$\pm$0.10          	\\
\quad bart-samsum 	& \max{0.26$\pm$0.11}	& \max{0.10$\pm$0.08}	& \max{0.17$\pm$0.08}	& \oosmark{-3.94$\pm$0.60}	& \oosmark{0.21$\pm$0.10}	\\
\quad distilbart  	& 0.22$\pm$0.10      	& \max{0.07$\pm$0.08}	& \max{0.15$\pm$0.08}	& \oosmark{-4.44$\pm$0.64}	& 0.17$\pm$0.10          	\\
\quad t5-small    	& 0.22$\pm$0.11      	& \max{0.07$\pm$0.08}	& \max{0.15$\pm$0.08}	& \oosmark{-4.62$\pm$0.62}	& 0.16$\pm$0.11          	\\

\bottomrule
\end{tabularx}
\caption{
Task A ``classical'' 
evaluation results (mean$\pm$std) on the ELMI and EuroParlMin datasets as summarized by 2023 systems (upper parts) and 2025 systems (lower parts).
  Systems are sorted by decreasing ROUGE-1.
  ``\oossymb'' denotes cells where the value increases, breaking the ordering. 
  Boldfaced scores are the best in each group, or within the $\pm$ bounds of the best scores.
}
\label{tab:task-a-automatic-eval}
\end{table*}

\section{Results}

This section presents the official results of AutoMin 2025.

\subsection{Task A Results}

We deem the LLM-based evaluation as the primary one in 2025.
\cref{tab:task-a-gpt-2025-cot} presents the scores obtained. We report the mean and standard deviation of all metric scores across the English test set meetings for both the ELITR Minuting (ELMI) and EuroParlMin corpora.
Note that due to the absence of contestants in Czech, we exclude the Czech version of ELMI in this report.

One important observation is that the Adequacy and Relevance scores are very similar, in many cases almost identical.
As in the previous instance, grammaticality is less discerning than the other measures because the outputs are overall gramatically good.
As previously noted by \citet{automin:2023}, GPT-based automatic scorers tend to
favor summaries produced by GPT-family models (notably GPT-4 and davinci-003),
likely due to alignment in style and phrasing.

Except for EuroParlMin in 2023, GPT-4 is again the best system of the tested ones. 
The low scores of the 2023 outputs can be attributed to the very short outputs in 2023: the average number of items that GPT-4 produced in the summary was only 5.0$\pm$0.6, much shorter than what all other systems produced.

Team \textit{HallucinationIndexes}'  submissions to AutoMin 2025 vary sharply
in quality across domains. On the ELMI~EN dataset, their summaries exhibit low
Adequacy and Relevance, suggesting a significant drop in content preservation.
However, Fluency (e.g. 2.92$\pm$0.51 for the bart-samsum model) and Grammar (3.50$\pm$0.67) remain reasonably
strong, indicating that while the output is well-formed linguistically, it
often fails to retain essential source information. In contrast, their system
performs markedly better on EuroParlMin, achieving top-tier Fluency
(4.41$\pm$0.68) and Grammar (4.63$\pm$0.76), with Adequacy
(2.96$\pm$0.99) and Relevance (3.52$\pm$1.16) outperforming or matching the
best systems from the 2023 instance. This contrast reflects the model's dependence
on surface alignment: it performs best when source--summary mappings are
tighter and structurally constrained.

\begin{figure}[t]
    \centering
    \includegraphics[width=\columnwidth]{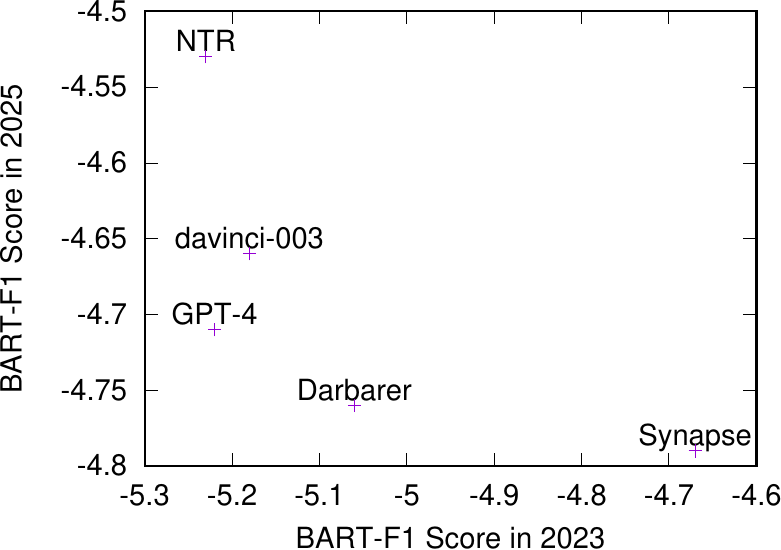}
    \caption{Illustration of BART-F1 score instability over years. We plot the scores of AutoMin 2023 systems on the EuroParlMin test set as reported in 2023 and calculated now. We see negatively correlated behaviour.}
    \label{bart-strange-result}
\end{figure}


The ``classical'' evaluation metrics are provided in \cref{tab:task-a-automatic-eval} for a comparison.
One immediate observation is that \textit{HallucinationIndexes} on ELMI EN fares much worse than the 2023 submissions in ROUGE, BART and BERT. Except for this substantially worse-scoring system, the classical metrics are unable to discern the best system from the rest with almost all scores falling into the same error bars.

\afterpage{\clearpage}

On the side, we mention that for a reliable comparison, we calculated the classical scores for both AutoMin rounds anew. Sadly, even the simple scores like ROUGE-1 in our \cref{tab:task-a-automatic-eval} occasionally differ from the ones reported in Table~7 in the 2023 overview paper. The reason for these differences lies in slight variations of UTF-8 handling and, e.g. emojis, present in output of some of the systems. Pearson correlations between the 2023 and 2025 calculations for each of ROUGE-* are mostly perfect (1.0) with three cases correlating above .97. The difference in BERT-F1 behaviour on EuroParlMin is larger, only .79. 
\label{bart-strange-disc}
BART-F1 is very unstable, with .63 correlation for ELMI-EN and \emph{negative} correlation of -.69 on EuroParlMin. The last case is depicted in \cref{bart-strange-result}.


\begin{table}[t]
\centering
\begin{tabular}{@{}l c@{}}
\toprule
\textbf{Approach} & \textbf{Mean Score} \\
\midrule
baseline\_gpt-4o-2024-11-20 & 7.74 \\
baseline\_llama-3.1-8B-instruct & 7.08 \\
baseline\_phi-4-mini-instruct & 6.84 \\
baseline\_phi-3-small-128k-instruct & 6.65 \\
baseline\_llama-3.2-3B-instruct & 6.33 \\
GETALP@AutoMin & 4.55 \\
GETALP@AutoMin\_amr & 4.34 \\
GETALP@AutoMin\_amr\_only & 2.73 \\
HallucinationIndexes@AutoMin & 2.28 \\
\bottomrule
\end{tabular}
\caption{Monolingual subtask (English-only) results for Task B. The reported performance corresponds to the average of scores within a 1--10 range (higher is better).}
\label{task-b-monoling-results}
\hfill
\end{table}

\subsection{Task B Results}
\label{sec:task-b-results}

The results of Task B are presented in \cref{task-b-monoling-results,task-b-crossling-results} for the monolingual and cross-lingual question-answering subtasks, respectively. The scores are based on evaluations by an LLM-judge using a scoring rubric from 1 to 10 (with higher scores indicating better responses) following the methodology described earlier. They represent the average over the scores obtained on the 130 questions of the \ebench{} test set.

For the monolingual subtask, our results confirm those of \citet{thonet2024elitrbenchmeetingassistantbenchmark}, who showed that LLMs such as GPT-4o and LLaMA-3.1/3.2 perform strongly on the meeting QA long-context task, with closed models still outperforming open ones. We also observe that the baseline based on Phi-4-Mini-Instruct obtained a highly competitive score, suggesting it to be a strong model in the 3-4B parameter class.

In contrast, the participants' submissions (\textit{GETALP} and \textit{HallucinationIndexes}) received significantly lower scores than our baselines, although \textit{GETALP} outperformed \textit{HallucinationIndexes}.
 We hypothesize that this is primarily due to the extractive nature of both systems, which may be disadvantaged by LLM-based evaluation methods. Moreover, extractive approaches are generally less effective than abstractive models when it comes to answering complex questions that require capturing nuanced information, understanding context, or performing reasoning.
In the specific case of \textit{HallucinationIndexes}, the lower performance is likely also due to limited handling of long-context inputs: the system is based on Flan-T5, which has a maximum context window of just 1,024 tokens, while meeting transcripts frequently exceed 10,000 tokens. If this limitation was not properly addressed (e.g., via chunking, summarization cascades, or retrieval mechanisms), it raises concerns about the reliability of the reported results, as the model would only process a small portion of the transcript, potentially overlooking key information necessary for accurate answers or summaries.


\begin{table}[t]
\centering
\begin{tabular}{@{}l c@{}}
\toprule
\textbf{Approach} & \textbf{Mean Score} \\
\midrule
baseline\_gpt-4o-2024-11-20 & 7.69 \\
baseline\_llama-3.1-8B-instruct & 6.21 \\
baseline\_phi-4-mini-instruct & 5.41 \\
baseline\_llama-3.2-3B-instruct & 5.11 \\
baseline\_phi-3-small-128k-instruct & 4.77 \\
GETALP@AutoMin\_amr & 3.11 \\
GETALP@AutoMin & 2.85 \\
GETALP@AutoMin\_amr\_only & 2.15 \\
\bottomrule
\end{tabular}
\caption{Cross-lingual subtask (English-Czech) results for Task B. The reported performance corresponds to the average of scores within a 0--10 range (higher is better).}
\label{task-b-crossling-results}
\end{table}

For the cross-lingual task, we observe that GPT-4o remains quite robust, achieving a score very close to that of the monolingual setting. In contrast, open models such as LLaMA and Phi experience a performance drop of more than one point in this more challenging cross-lingual scenario.
Among the participants, only \textit{GETALP}
 submitted a system for the cross-lingual task, and we similarly observe a notable degradation in performance compared to their monolingual results.



\section{Discussion on Task A Evaluation}

The two years between AutoMin 2023 and the current AutoMin 2025 saw a big improvement in LLM performance and reliability. It is thus very interesting to compare how GPT evaluated AutoMin 2023 systems in 2023 and how it is evaluating them (and also the new 2025 submission) now.

\begin{table*}[t]
    \centering
    \small
    \begin{tabular}{lcccc|cccc}
& \multicolumn{4}{c|}{GPT-2023-AFGR [0,10]} & \multicolumn{4}{c}{GPT-2023-AFGR rescaled to [1,5]}  \\
& Adeq.& Flu.& Gram. & Relev. & Adeq.& Flu. & Gram. & Relev.
\\
\hline\multicolumn{8}{l}{ELMI EN}\\ \hline
GPT-4       &\max{8.75$\pm$0.45}&\max{8.83$\pm$0.39}          &\max{9.00$\pm$0.00}    &\max{8.75$\pm$0.45}	&\max{4.50$\pm$1.18}&\max{4.53$\pm$1.16}          &\max{4.60$\pm$1.00}    &\max{4.50$\pm$1.18}	\\
davinci-003 &8.00$\pm$0.85      &\max{8.58$\pm$0.67}          &8.83$\pm$0.58          &8.00$\pm$0.85      	&4.20$\pm$1.34      &\max{4.43$\pm$1.27}          &4.53$\pm$1.23          &4.20$\pm$1.34      	\\
Zoom-long   &7.83$\pm$0.39      &8.42$\pm$0.51                &8.75$\pm$0.45          &7.83$\pm$0.39      	&4.13$\pm$1.16      &4.37$\pm$1.20                &4.50$\pm$1.18          &4.13$\pm$1.16      	\\
Darbarer    &7.58$\pm$0.67      &\max{\oosmark{8.50$\pm$0.67}}&\oosmark{8.83$\pm$0.39}&7.58$\pm$0.67      	&4.03$\pm$1.27      &\max{\oosmark{4.40$\pm$1.27}}&\oosmark{4.53$\pm$1.16}&4.03$\pm$1.27      	\\
Synapse     &7.42$\pm$0.90      &8.25$\pm$0.75                &8.58$\pm$0.67          &7.42$\pm$0.79      	&3.97$\pm$1.36      &4.30$\pm$1.30                &4.43$\pm$1.27          &3.97$\pm$1.32      	\\
NTR         &7.08$\pm$0.90      &7.83$\pm$0.72                &8.08$\pm$0.67          &7.25$\pm$1.14      	&3.83$\pm$1.36      &4.13$\pm$1.29                &4.23$\pm$1.27          &3.90$\pm$1.46      	\\
Team Iterate&6.58$\pm$1.38      &7.67$\pm$0.98                &\oosmark{8.17$\pm$0.72}&6.75$\pm$1.22      	&3.63$\pm$1.55      &4.07$\pm$1.39                &\oosmark{4.27$\pm$1.29}&3.70$\pm$1.49      	\\
    \end{tabular}
    \caption{Task A results from 2023, reproduced from Table~8 of
	\perscite{automin:2023}: Automatic evaluation results using GPT
	in 2023 using the ``AFGR'' prompt, see \cref{gpt2023-prompt}.
    \\
    The left part of the table reports the scores exactly as in \perscite{automin:2023}, in the range 0--10. The right part of the table rescales them linearly to fit the current range 1--5 (both worst-to-best).
    \\
    We report
	the average $\pm$ standard deviation. Sorted by decreasing Adequacy estimated by GPT-AFRG.
    The symbol ``\oossymb{}'' highlights a disruption in the ordering in the given column.
The top score and all scores that fall within its
std. dev. bounds are in \textbf{bold}.
	}
    \label{tab:gpt2023-afgr}
\end{table*}

In addition to the improvements that happened within GPT-4, we also improved our prompt.
For our analysis, we look at all available combinations:

\begin{description}
\item[GPT-2025-CoT] denotes the evaluation collected this year with the current
GPT and the Chain-of-Thought prompt (\cref{gpt2025-prompt} in the Appendix). These are the scores we provided in \cref{tab:task-a-gpt-2025-cot}.
\item[GPT-2023-AFGR] denotes the evaluation collected in 2023 with GPT back then and the simple prompt asking for Adequacy, Fluency, Grammaticality and Relevance in one go, see \cref{gpt2023-prompt} in the Appendix. These scores are reproduced in \cref{tab:gpt2023-afgr}.
\item[GPT-2025-AFGR] is a combination: the current GPT using the simple AFGR prompt (\cref{gpt2023-prompt}), see the results in \cref{tab:task-a-gpt-2025-afrg} in the Appendix.
\end{description}

\label{afgr-rescaling}
A very important caveat is that the AFGR prompt asks the model to use the scale from 0 (worst) to 10 (best), whereas the CoT prompts asks for a smaller scale from 1 to 5. We \emph{keep} the AFGR prompt unchanged and only afterwards, once the scores are obtained, we rescale them using the formula $y=\frac{4}{10}x+1$ for a better comparability with the current scores.

\begin{figure*}[ht]
  \centering
  \includegraphics[width=\textwidth]{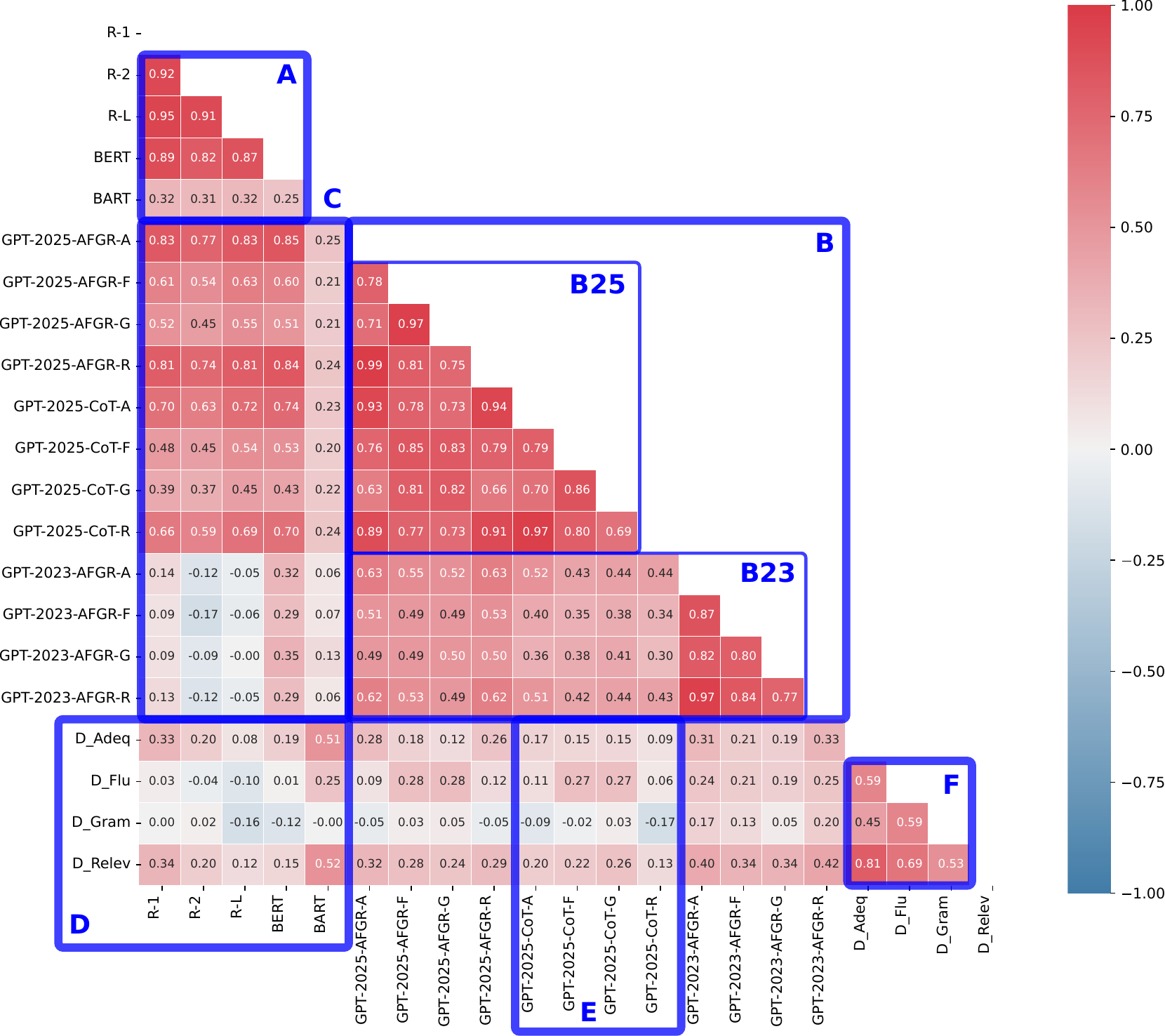}
  \caption{Pearson correlations of different metrics on ELMI EN dataset
  calculated at the meeting level on outputs from both 2023 and 2025 systems.
  GPT-2023-AFGR scores are used after the rescaling described in
  \cref{afgr-rescaling}.}
  \label{fig:heatmap}
\end{figure*}

\subsection{Pearson Correlation of Metrics}

\cref{fig:heatmap} provides Pearson correlations of all the examined scoring
methods on the ELMI EN minutes as produced by both 2023 systems and 2025
systems. The correlations are, as in 2023, calculated using individual meetings as the
datapoints and considering all applicable systems' outputs for the given
evaluation method. (In particular, we cannot obtain GPT-2023 scores for 2025
outputs.)

In the top left corner, highlighted as region A, we see that ROUGE variants
(R-1, R-2, and R-L) are
correlated very closely with each other, and BERT is also very close, correlating
at $\sim$0.89 with ROUGE-1. 
BART differs, as we could expect from its strange change between the years
discussed in \cref{bart-strange-disc} above.

We now turn our attention to GPT-based evaluation, see the region B. All
GPT-based scores are quite correlated with one another.
This hold esp. for all scores obtained in the 2025 runs (region B25) where the
lowest correlation is 0.63 between 2025 CoT-Grammaticality and AFGR-Adequacy, a
score that one would really expect low.

As we see in the region B23, GPT-2023-* does not agree too well with GPT-2025-* scores, with Pearson of at most
0.63.
The only high correlations 
we observe for GPT-2023-* are the mutually correlated assessments where GPT 2023 saw Adequacy not only
correlated to Relevance (at 0.97 Pearson) but also, e.g., to Fluency (at 0.87 Pearson),
see the right three columns in the region B23.

The correlation between the classical evaluation methods and GPT-based evaluation
varies across the years, see the region C in the heatmap. For GPT-2025-* scores,
the correlations reach
up to 0.85 Pearson.
The scores obtained by GPT-4 in 2023 were however different and we
rightfully criticized them in \perscite{automin:2023}. The correlation of
GPT-2023-* and classical measures (bottom 4 rows in the region C) are typically at the
level of 0.1 with BERT reaching 0.3.

The last four rows relate our automatic methods to the manual
judgements collected in 2023, also at the level of whole meetings
(``document-level'', thus denoted ``D\_'').

The correlations between the classical evaluation methods and the manual
annotations (region D in the bottom-left corner) suggest that one should really
search for better metrics. It is reasonable to expect these essentially word-level measures not to
correlate with Grammaticality (D\_Gram) and Fluency (D\_Flu) but they are
supposed to approximate Adequacy and Relevance. BART is the best, at 0.5
Pearson, followed by ROUGE-1
with 0.3 and other correlations even lower.

Sadly, the GPT-style evaluation, and esp. the one that we intended to make our
official scoring method (GPT-2025-CoT) performs worse than ROUGE-1: Pearson is
at .17 or .13 for Adequacy and Relevance, resp., see the region E.

Finally, we discuss the Adequacy and Relevance scales.
The 2025 evaluations, both for AFGR and CoT prompts, document that these two
scales are regarded as almost identical, see e.g. the Pearson of 0.99 between
GPT-2025-AFGR-A and GPT-2025-AFGR-R and similarly (.97) between GPT-2025-CoT-A and
GPT-2025-CoT-R, all in the region B. The 2023 GPT-evaluation is not different,
with GPT-2023-AFGR-A correlating GPT-2023-AFGR-R at 0.97.
We assume that the terms ``adequacy'' and ``relevance'' can be seen as near synonyms in the general
language. Our definitions try to draw a distinction:

\begin{itemize}
\item Adequacy considers
the coverage of important topics, i.e. serves as a recall-type measure: ``Do the
minutes summarize all important things?''
\item Relevance definition highlights that the score should sink when useless parts are summarized, i.e. it should serve as a precision-type measure.
\end{itemize}

While we are not sure if GPT was able to reflect this slight difference, it is
conceivable that the minutes are comparable in these regards. We are thus not
very surprised to see precision-like Relevance and recall-like Adequacy
correlated across the participating systems. This is also confirmed by the 2023
manual judgements (region F in \cref{fig:heatmap}), where the
Adequacy--Relevance correlation is the highest one, 0.81.

We can also discuss the relation between the relatively straightforward AFGR prompt vs. the Chain-of-Thought prompt introduced this year.
In both cases, GPT was expected to produce all four scores in one prediction.
This strategy was problematic in 2023 where all four AFGR scores were highly
correlated (region B23 right) but it works better this year (region B25). In 2025, looking e.g. at GPT-2025-CoT-G, we
see that it correlates very high (0.86) only with GPT-2025-CoT-F while the
correlations with Adequacy and Relevance stay somewhat lower, at the 0.7 level.

Focusing on the question whether CoT is better than AFGR according to
manual evaluation, we have to conclude that the correlations of 2025-CoT (region
E) are actually lower than for AFGR as obtained in 2025 (left of the region E) and
even more for AFGR obtained in 2023 (right of the region E). The switch to
Chain-of-Though thus did not bring any useful improvement.

Given the multiple discrepancies in these evaluations, we should plan a new
round of manual assessments for the future.

\subsection{A Small Adversarial Case Study}
\label{adversarial-study}

When processing our data, we noticed at one stage that GPT-4 outputs from 2023
were messed up for the EuroParlMin dataset, attached to a wrong meeting. We took
this opportunity and reviewed 91 meetings manually, locating 45 misaligned
minutes. These minutes can be seen as adversarial inputs for our evaluation
metrics: Because the minutes are not related to the meeting, they should score
much lower than other meetings.

\def\signdiff{\cmark}
\def\nosigndiff{\xmark}
\begin{table*}[t]
\begin{center}
\begin{tabular}{lrrrl}
               	& Adversarial Minutes	& Regular Minutes	& $\Delta$ & Sign. Diff \\
\hline
GPT-2025-CoT-R 	& 1.24$\pm$0.77 	& 4.68$\pm$0.61 	&     3.44	& \signdiff{} (p=0.0000) \\
GPT-2025-CoT-A 	& 1.20$\pm$0.59 	& 4.61$\pm$0.62 	&     3.41	& \signdiff{} (p=0.0000) \\
GPT-2025-AFGR-R	& 1.37$\pm$0.54 	& 4.19$\pm$0.60 	&     2.82	& \signdiff{} (p=0.0000) \\
GPT-2025-AFGR-A	& 1.43$\pm$0.51 	& 4.22$\pm$0.60 	&     2.79	& \signdiff{} (p=0.0000) \\
BART           	& -4.75$\pm$0.61	& -3.97$\pm$0.49	&     0.78	& \signdiff{} (p=0.0000) \\
GPT-2025-AFGR-F	& 3.99$\pm$0.72 	& 4.64$\pm$0.32 	&     0.65	& \signdiff{} (p=0.0000) \\
GPT-2025-CoT-F 	& 4.64$\pm$0.91 	& 4.93$\pm$0.25 	&     0.29	& \signdiff{} (p=0.0071) \\
GPT-2025-AFGR-G	& 4.72$\pm$0.88 	& 4.95$\pm$0.14 	&     0.23	& \nosigndiff{} (p=0.2631) \\
GPT-2025-CoT-G 	& 4.82$\pm$0.83 	& 4.98$\pm$0.15 	&     0.16	& \nosigndiff{} (p=0.4069) \\
BERT           	& 0.12$\pm$0.12 	& 0.26$\pm$0.13 	&     0.14	& \signdiff{} (p=0.0000) \\
R-1            	& 0.17$\pm$0.06 	& 0.30$\pm$0.13 	&     0.13	& \signdiff{} (p=0.0000) \\
R-2            	& 0.03$\pm$0.02 	& 0.12$\pm$0.08 	&     0.09	& \signdiff{} (p=0.0000) \\
R-L            	& 0.11$\pm$0.04 	& 0.19$\pm$0.09 	&     0.08	& \signdiff{} (p=0.0000) \\

\end{tabular}
\end{center}
\caption{Average metric scores ($\pm$standard deviations) for GPT-4-2023 and
GPT-4-2025 minutes of 91 EuroParlMin meetings. 45 of the GPT-4-2023 outputs are
adversary in that they belong to another meeting. The average scores in the
``Adversarial Minutes'' should be much lower than in ``Regular Minutes''.
$\Delta$ reports the absolute difference and serves as the basis for sorting.
The last column reports the statistical significance of the difference according
to Mann-Whitney test.
}
\label{adversarial-summary}
\end{table*}

\cref{adversarial-summary} confirms that almost all metrics are able to identify
unrelated minutes by giving them a much lower score. Of the LLM-based
evaluations, only Grammaticality does not allow us to separate the adversarial
and regular minutes. Fluency, esp. in the AFGR prompt, is obviously affected by
other scores (Relevance and Adequacy) that the LLM is assigning to the given
candidate, so even Fluency scores are significantly different for matching and
mismatching minutes, although GPT-2025-CoT Fluency has at least a little higher
p-value.

All the classical metrics are also robust to this type of adversarial attach,
i.e. the diverging set of words appearing in the transcript and minutes is
sufficient to identify the mismatch.

Looking at the GPT scores, all of which are in the 1--5 range, we can also look
at the absolute difference of scores in these groups.
Here we see that the CoT
version of the prompt leads to a higher difference, so it could be deemed more 
robust in evaluation of adversarial summaries than the AFGR prompt.

\section{Conclusion}

This paper presented AutoMin 2025, a shared task on automatic meeting summarization into minutes, as well as question answering (QA) based on meeting transcripts.

Participation in 2025 was limited compared to previous years, raising important questions about the evolving interest in meeting processing and summarization particularly in light of the rapid progress and widespread use of general-purpose large language models (LLMs). To support meaningful evaluation despite the lower participation, we included several baseline systems and top-performing submissions from the 2023 edition. This allows for a comprehensive assessment of current-generation LLMs on both tasks.

Notably, the new QA task offers a valuable benchmark for evaluating long-context LLMs, as our 1-hour meeting transcripts required processing context windows exceeding 16k tokens.

In the traditional minuting, Task A, we confirmed the preference GPT-based
evaluation is giving to GPT outputs. We experimented with a Chain-of-Thought
prompt in the evaluation, hoping to improve the reliability of the scorings. The
2025 outputs of GPT-4 indicate that both the straighforward AFGR prompt as the
more complex CoT prompt give similar outputs, different from what we obtained
using AFGR in 2023. In our case, the CoT prompting provides hardly any
improvement over the simple AFGR prompt. On the negative side, it correlates
with human judgements a little less than AFGR. On the positive side, it is a little more robust to
adversarial inputs and the score components (e.g. Relevance vs. Grammaticality)
seem to be estimated more independently of each other because in our small
adversarial case study, Grammaticality rightfully did not allow to identify
minutes belonging to a different meeting.

We hope the scientific community will find our datasets and evaluation metrics useful for advancing research in meeting processing and for developing more capable systems in this domain.

\section*{Acknowledgments}
This paper was partially funded by the European Commission through the UTTER project under grant number
101070631.

This work has also received funding from the Project OP JAK Mezisektorová spolupráce Nr. CZ.02.01.01/00/23\_020/0008518 named ``Jazykověda, umělá inteligence a jazykové a řečové technologie: od výzkumu k aplikacím.''

\bibliography{biblio,main-elitr}

\clearpage
\onecolumn
\appendix
\section{Appendix}

\begin{center}
\begin{tcolorbox}[colback=white, colframe=black, width=0.95\textwidth, sharp corners=south, boxrule=0.5pt]
\small
\textbf{PROMPT:}

You are an expert meeting‑summary evaluator. Given a meeting transcript and its corresponding meeting minutes, your task is to score the minutes on four criteria—\textbf{adequacy}, \textbf{fluency}, \textbf{grammaticality}, and \textbf{relevance}—using a 1–5 scale (1 = very poor, 5 = excellent). 

Use these paper‑sourced definitions exactly:

\begin{itemize}
  \item \textbf{Adequacy}: “Rates the amount of meaning expressed in the generated sample given a reference sample.”  
  \item \textbf{Fluency}: “Represents the quality of expression.”  
  \item \textbf{Grammaticality}: “Measures the capability of a model to produce grammatically correct texts. [Assessed by counting types of errors.]”  
  \item \textbf{Relevance}: “Represents how closely are the documents related.”  
\end{itemize}

\textbf{Instructions:}
\begin{enumerate}
  \item \textbf{Step‑by‑Step Analysis}: Before assigning scores, list all major points in the minutes and cite the corresponding transcript segments.  
  \item \textbf{Metric‑by‑Metric Evaluation}: For each metric, briefly \textbf{explain} (1–2 sentences) why you chose that score, referencing the definitions.  
  \item \textbf{Avoid Vagueness}: Tie your comments to concrete examples (e.g., “Sentence 3 of minute 2 uses ‘…’ which is ungrammatical because…”).  
  \item \textbf{Self‑Check}: After scoring, re‑evaluate your lowest score and confirm it cannot be higher based on the definitions.  
  \item \textbf{Final Output Only}: Return \textbf{strictly} the JSON object defined below—no extra commentary.
\end{enumerate}

\textbf{Output JSON Schema}
\begin{verbatim}
{
  "scores": {
    "adequacy": <int 1–5>,
    "fluency": <int 1–5>,
    "grammaticality": <int 1–5>,
    "relevance": <int 1–5>
  },
  "rationale": {
    "adequacy": "<brief justification>",
    "fluency": "<brief justification>",
    "grammaticality": "<brief justification>",
    "relevance": "<brief justification>"
  }
}
\end{verbatim}

Now evaluate: \\
Transcript: \textcolor{blue}{\{transcript\}} \\
Minutes: \textcolor{blue}{\{minutes\}} 
\end{tcolorbox}
\end{center}
\captionof{figure}{Task A evaluation prompt used this year, called GPT-2025-CoT. The transcript and minutes were inserted as plain text at the indicated places in the prompt. The text formatting (except for font face), i.e. indentation and boldfacing, were available to GPT, too, in the common markdown style, e.g.: **Adequacy**: “Rates the amount\dots”}
\label{gpt2025-prompt}

\begin{figure*}[t]
  \centering
\begin{tcolorbox}[colback=white, colframe=black, width=0.95\textwidth, sharp corners=south, boxrule=0.5pt]
    \parbox{.9\textwidth}{%
        \#\#\# Task description:\\
        You are provided below with a question, a response to evaluate, a reference answer that gets the maximum score of 10, and a score rubric representing evaluation criteria.\\
        1. Write a detailed feedback that assess the quality of the response strictly based on the given score rubric, not evaluating in general.\\
        2. After writing a feedback, write a score that is an integer between 1 and 10. You should refer to the score rubric.\\
        3. The output format should first include the feedback and then indicate the integer score in \textbackslash boxed\{\}.\\
        4. Please do not generate any other opening, closing, and explanations.\\\\
        \#\#\# Question:\\
        \textcolor{blue}{\{question\}}\\\\
        \#\#\# Response to evaluate:\\
        \textcolor{blue}{\{response\}}\\\\
        \#\#\# Reference answer (score 10):\\
        \textcolor{blue}{\{reference\}}\\\\
        \#\#\# Score rubric:\\
        {[Does the response to evaluate correctly address the given question based on the elements provided by the reference answer? The response should include the elements of the reference answer and should also avoid adding unnecessary elements or being too verbose.]}\\
        Score 1: The response to evaluate is incorrect and misses all the elements of the reference answer.\\
        Score 2: The response to evaluate indicates insufficient knowledge to answer the question even though the reference answer states otherwise.\\
        Score 3-4: The response to evaluate contains some elements vaguely related to the reference answer.\\
        Score 5-6: The response to evaluate is partially correct and/or covers only a part of the reference answer.\\
        Score 7-8: The response to evaluate contains most of the reference answer but delivers it in an indirect and/or overly verbose way.\\
        Score 9: The response to evaluate includes the reference answer but it is more verbose and adds unnecessary elements.\\
        Score 10: The response to evaluate is essentially equivalent to the reference answer.\\\\
        \#\#\# Feedback:
        \\
    }
\end{tcolorbox}
  \caption{Evaluation prompt for the LLM-as-a-judge evaluator used in the monolingual (English-only) subtask of Task B, based on~\citet{thonet2024elitrbenchmeetingassistantbenchmark}. The elements in blue and enclosed in curly brackets correspond to question-specific text spans that are dynamically adapted.}
  \label{fig:task-b-prompt-mono}
\end{figure*}

\begin{figure*}[t]
  \centering
\begin{tcolorbox}[colback=white, colframe=black, width=0.95\textwidth, sharp corners=south, boxrule=0.5pt]
    \parbox{.9\textwidth}{%
        \#\#\# Task description:\\
        You are provided below with a question, a response to evaluate, a reference answer that gets the maximum score of 10, and a score rubric representing evaluation criteria.\\
        The question and the reference answer are in Czech.\\
        1. Write a detailed feedback that assess the quality of the response strictly based on the given score rubric, not evaluating in general.\\
        2. After writing a feedback, write a score that is an integer between 0 and 10. You should refer to the score rubric.\\
        3. The output format should first include the feedback and then indicate the integer score in \textbackslash boxed\{\}.\\
        4. Please do not generate any other opening, closing, and explanations.\\\\
        \#\#\# Question:\\
        \textcolor{blue}{\{question\}}\\\\
        \#\#\# Response to evaluate:\\
        \textcolor{blue}{\{response\}}\\\\
        \#\#\# Reference answer (score 10):\\
        \textcolor{blue}{\{reference\}}\\\\
        \#\#\# Score rubric:\\
        {[Does the response to evaluate correctly address the given question based on the elements provided by the reference answer? The response should be in Czech, it should include the elements of the reference answer and should also avoid adding unnecessary elements or being too verbose.]}\\
        Score 0: The response to evaluate is not in Czech language.\\
        Score 1: The response to evaluate is incorrect and misses all the elements of the reference answer.\\
        Score 2: The response to evaluate indicates insufficient knowledge to answer the question even though the reference answer states otherwise.\\
        Score 3-4: The response to evaluate contains some elements vaguely related to the reference answer.\\
        Score 5-6: The response to evaluate is partially correct and/or covers only a part of the reference answer.\\
        Score 7-8: The response to evaluate contains most of the reference answer but delivers it in an indirect and/or overly verbose way.\\
        Score 9: The response to evaluate includes the reference answer but it is more verbose and adds unnecessary elements.\\
        Score 10: The response to evaluate is essentially equivalent to the reference answer.\\\\
        \#\#\# Feedback:
        \\
    }
\end{tcolorbox}
  \caption{Evaluation prompt for the LLM-as-a-judge evaluator used in the cross-lingual (English-Czech) subtask of Task B, based on~\citet{thonet2024elitrbenchmeetingassistantbenchmark}. The elements in blue and enclosed in curly brackets correspond to question-specific text spans that are dynamically adapted.}
  \label{fig:task-b-prompt-cross}
\end{figure*}

\begin{figure*}[t]
  \centering
\begin{tcolorbox}[colback=white, colframe=black, width=0.95\textwidth, sharp corners=south, boxrule=0.5pt]
    \parbox{12cm}{%
        The following is the transcript of a meeting with multiple participants, where utterances start with the speaker's anonymized name (for instance (PERSON4)) and may span over several lines.\\\\
        \textcolor{blue}{\{transcript\}}\\\\
        As a professional conversational assistant, your task is to answer questions about the meeting by making inferences from the provided transcript.\\\\
        \textcolor{blue}{\{question\}}
        \\
    }
\end{tcolorbox}
  \caption{Answer prompt used to obtain responses from the prompted LLM baselines on the monolingual (English-only) subtask of Task B, based on~\citet{thonet2024elitrbenchmeetingassistantbenchmark}. The elements in blue and enclosed in curly brackets correspond to meeting-specific and question-specific text spans that are dynamically adapted.}
  \label{fig:answer-prompt-en}
\end{figure*}

\begin{figure*}[t]
  \centering
\begin{tcolorbox}[colback=white, colframe=black, width=0.95\textwidth, sharp corners=south, boxrule=0.5pt]
    \parbox{12cm}{%
        The following is the transcript of a meeting with multiple participants, where utterances start with the speaker's anonymized name (for instance (PERSON4)) and may span over several lines.\\\\
        \textcolor{blue}{\{transcript\}}\\\\
        As a professional conversational assistant, your task is to answer questions about the meeting by making inferences from the provided transcript. Although the transcript is in English, the questions are in Czech and the answers should also be formulated in Czech.\\\\
        \textcolor{blue}{\{question\}}
        \\
    }
\end{tcolorbox}
  \caption{Answer prompt used to obtain responses from the prompted LLM baselines on the cross-lingual (English-Czech) subtask of Task B, based on~\citet{thonet2024elitrbenchmeetingassistantbenchmark}. The elements in blue and enclosed in curly brackets correspond to meeting-specific and question-specific text spans that are dynamically adapted.}
  \label{fig:answer-prompt-encz}
\end{figure*}

\begin{figure*}
\centering
\begin{tcolorbox}[colback=white, colframe=black, width=0.95\textwidth, sharp corners=south, boxrule=0.5pt]
    \parbox{.9\textwidth}{%
Given the following meeting transcript and minutes, evaluate the minutes for their adequacy (to what extent the minutes adequately capture the major topics discussed in the meeting, also considering coverage, i.e. all such topics covered), fluency (if the minutes consist of fluent, coherent texts and are readable to the evaluator), grammatical correctness (the level to which the minutes are grammatically correct) and relevance (the extent to which the minutes overall capture the important content from the source transcript (as opposed to summarizing useless parts). \\
\_\_\_\_\_\_\_\_\_\_\_\_\_\_\_\_\_\_\_\_\_\_\_\_\_\_\_\_\_\_\_\_\_\_\_\_\_\_\_\_\_\_\_\_\_\_\_\_\_\_\_\_\_\_\_\_\_\_\_\\
Transcript:\\
\textcolor{blue}{\{transcript\}} \\
\_\_\_\_\_\_\_\_\_\_\_\_\_\_\_\_\_\_\_\_\_\_\_\_\_\_\_\_\_\_\_\_\_\_\_\_\_\_\_\_\_\_\_\_\_\_\_\_\_\_\_\_\_\_\_\_\_\_\_\\
Minutes:\\
\textcolor{blue}{\{system\_generated\_minutes\}}\\
\_\_\_\_\_\_\_\_\_\_\_\_\_\_\_\_\_\_\_\_\_\_\_\_\_\_\_\_\_\_\_\_\_\_\_\_\_\_\_\_\_\_\_\_\_\_\_\_\_\_\_\_\_\_\_\_\_\_\_\\
Now evaluate the minutes for their adequacy, fluency, grammatical correctness and relevance. Give each score separately on a scale 0 to 10, where 10 is the best: \\
    }
\end{tcolorbox}
\caption{Task A evaluation prompt used in 2023, called ``GPT-AFGR'' in \citet{automin:2021} and reported two times this year as GPT-2023-AFGR and GPT-2025-AFGR. We rescale the scores obtained as described in \cref{afgr-rescaling}.}
\label{gpt2023-prompt}
\end{figure*}

\begin{table*}[t]
  \centering
  \begin{tabular*}{\textwidth}{@{\extracolsep{\fill}} l c c c c }
	& \multicolumn{4}{c}{\bf GPT-2025-AFGR} \\
    & \textbf{Adeq.} & \textbf{Flu.} & \textbf{Gram.} & \textbf{Relev.} \\
    \cmidrule{1-5}
    \multicolumn{5}{@{}l}{\textbf{ELMI EN}} \\
    \cmidrule{1-5}
GPT-4           	& \max{4.30$\pm$0.18}	& \max{4.53$\pm$0.16}    	& \max{4.57$\pm$0.12}    	& \max{4.57$\pm$0.12} \\
text-davinci-003	& 3.77$\pm$0.12      	& 4.17$\pm$0.12          	& 4.27$\pm$0.23          	& 4.00$\pm$0.27 \\
davinci-003     	& 3.77$\pm$0.27      	& 4.07$\pm$0.20          	& 4.20$\pm$0.17          	& \oosmark{4.10$\pm$0.30} \\
Zoom-long       	& 3.20$\pm$0.36      	& 3.57$\pm$0.27          	& 3.50$\pm$0.35          	& 3.27$\pm$0.36 \\
Synapse         	& 3.07$\pm$0.23      	& 3.57$\pm$0.43          	& 3.43$\pm$0.60          	& 3.13$\pm$0.26 \\
Darbarer        	& 3.03$\pm$0.36      	& \oosmark{3.87$\pm$0.33}	& \oosmark{3.90$\pm$0.35}	& \oosmark{3.23$\pm$0.40} \\
\kmjec{}        	& 2.77$\pm$0.50      	& 3.03$\pm$0.47          	& 2.90$\pm$0.57          	& 2.80$\pm$0.53 \\
NTR             	& 2.57$\pm$0.21      	& 2.60$\pm$0.34          	& 2.40$\pm$0.21          	& 2.57$\pm$0.21 \\
Zoom-short      	& 2.30$\pm$0.30      	& \oosmark{3.50$\pm$0.39}	& \oosmark{3.63$\pm$0.67}	& 2.53$\pm$0.41 \\
    \cdashline{1-5}
GPT-4 (2025)                	& 3.97$\pm$0.43	& \max{4.57$\pm$0.21}    	& \max{4.70$\pm$0.18}    	& 4.30$\pm$0.39 \\
HallucinationIndexes@ (2025)	\\             	                         	                         	
\quad bart-samsum           	& 1.47$\pm$0.16	& 3.10$\pm$0.73          	& 3.20$\pm$0.86          	& 1.53$\pm$0.20 \\
\quad distilbart            	& 1.40$\pm$0.00	& 2.17$\pm$0.50          	& 2.13$\pm$0.29          	& 1.43$\pm$0.12 \\
\quad t5-small              	& 1.40$\pm$0.00	& 2.00$\pm$0.43          	& 2.00$\pm$0.43          	& 1.40$\pm$0.00 \\
\quad pegasus-xsum          	& 1.10$\pm$0.18	& \oosmark{2.60$\pm$1.07}	& \oosmark{2.93$\pm$1.17}	& 1.10$\pm$0.18 \\
    \cmidrule{1-5}
    \multicolumn{5}{@{}l}{\textbf{EuroParlMin}} \\
    \cmidrule{1-5}
Team Synapse	& 3.68$\pm$0.57	& 4.53$\pm$0.39          	& 4.79$\pm$0.40          	& 4.01$\pm$0.60 \\
NTR         	& 3.08$\pm$0.87	& 3.72$\pm$0.91          	& 3.93$\pm$0.96          	& 3.26$\pm$0.97 \\
Darbarer    	& 2.73$\pm$0.66	& \oosmark{4.25$\pm$0.36}	& \oosmark{4.61$\pm$0.44}	& 2.99$\pm$0.80 \\
davinci-003 	& 1.85$\pm$1.05	& 4.13$\pm$0.39          	& 4.51$\pm$0.36          	& 1.90$\pm$1.14 \\
GPT-4       	& 1.82$\pm$1.02	& \oosmark{4.23$\pm$0.43}	& \oosmark{4.59$\pm$0.41}	& 1.89$\pm$1.12 \\
    \cdashline{1-5}
GPT-4 (2025)                	& \max{4.61$\pm$0.47}	& \max{4.92$\pm$0.27}    	& \max{4.98$\pm$0.22}    	& \max{4.74$\pm$0.40} \\
HallucinationIndexes@ (2025)	\\                   	                         	                         	
\quad bart-samsum           	& 3.10$\pm$0.96      	& 4.28$\pm$0.53          	& 4.50$\pm$0.65          	& 3.36$\pm$1.04 \\
\quad distilbart            	& 2.86$\pm$0.87      	& 3.83$\pm$0.64          	& 4.04$\pm$0.83          	& 3.11$\pm$0.95 \\
\quad t5-small              	& 2.81$\pm$0.85      	& 3.29$\pm$0.59          	& 3.21$\pm$0.70          	& 3.07$\pm$0.97 \\
\quad pegasus-xsum          	& 1.89$\pm$0.57      	& \oosmark{3.48$\pm$0.87}	& \oosmark{3.83$\pm$0.96}	& 1.96$\pm$0.68 \\
    \bottomrule
  \end{tabular*}
  \caption{GPT-2025-AFGR evaluation results (mean$\pm$std) on ELMI EN and
  EuroParlMin datasets, with 2023 and 2025 systems. Scores rescaled to 1
  (worst)–5 (best). Systems sorted by Adequacy in each block. ``\oossymb'' marks out-of-order cells. Bold highlights the best in each block.}
  \label{tab:task-a-gpt-2025-afrg}
\end{table*}

\end{document}